\pdfoutput=1

\documentclass[11pt]{article}

\PassOptionsToPackage{dvipsnames}{xcolor}
\usepackage[preprint]{acl}

\usepackage{times}
\usepackage{latexsym}

\usepackage[T1]{fontenc}

\usepackage[utf8]{inputenc}

\usepackage{microtype}

\usepackage{inconsolata}

\usepackage{inconsolata}
\usepackage{geometry}
\usepackage{subcaption}
\usepackage{tabularx}
\usepackage{booktabs}
\usepackage{graphicx}
\usepackage{mdframed}
\usepackage{multirow}
\usepackage{amsmath}

\title{QA-prompting: Improving Summarization with Large Language Models using Question-Answering}

\author{
\textbf{Neelabh Sinha}$^{1}$ \\
$^1$Georgia Institute of Technology \\
{\tt neelabhsinha97@gmail.com}
}

\begin{document}
\maketitle
\begin{abstract}
Language Models (LMs) have revolutionized natural language processing, enabling high-quality text generation through prompting and in-context learning. However, models often struggle with long-context summarization due to positional biases, leading to suboptimal extraction of critical information. There are techniques to improve this with fine-tuning, pipelining, or using complex techniques, which have their own challenges. To solve these challenges, we propose \textit{QA-prompting} -- a simple prompting method for summarization that utilizes question-answering as an intermediate step prior to summary generation. Our method extracts key information and enriches the context of text to mitigate positional biases and improve summarization in a single LM call per task without requiring fine-tuning or pipelining. Experiments on multiple datasets belonging to different domains using ten state-of-the-art pre-trained models demonstrate that QA-prompting outperforms baseline and other state-of-the-art methods, achieving up to 29\% improvement in ROUGE scores. This provides an effective and scalable solution for summarization and highlights the importance of domain-specific question selection for optimal performance~\footnote{GitHub repository link of the implementation: \href{https://github.com/neelabhsinha/qa-prompting}{https://github.com/neelabhsinha/qa-prompting}}.
\end{abstract}


\section{Introduction}
\label{sec:intro}

Language Models (LMs) have revolutionized the application of Natural Language Processing. With instruction tuning~\citep{instructGPT}, prompting~\citep{gpt3}, and in-context learning~\citep{icl,icl_survey}, LLMs perform well in most of the conditional generation tasks out-of-the-box. Specifically, in abstractive summarization, this approach yields highly fluent, consistent, and relevant summaries~\citep{news_summarization} that are even preferred over summaries generated by state-of-the-art (SOTA) models like BRIO~\citep{brio}.

\begin{figure}
    \centering
    \includegraphics[width=\linewidth]{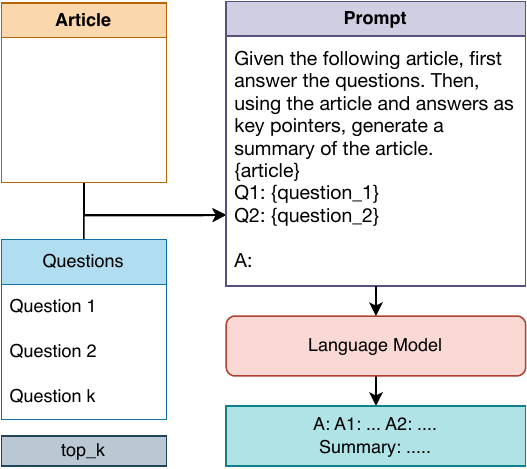}
    \caption{Framework for QA-prompting: Relevant questions are extracted from a corpus based on the domain of article, and a user defined top-$k$ value. A prompt is constructed to first answer the questions, followed by generating summary.}
    \label{fig:method_summary}
\end{figure}

Despite the metrics rating the summaries low, several works argue that LM summaries are of high quality~\citep{benchmarking_llm_summarization,sinha-etal-2025-small}, fluent, grammatically correct, and largely relevant. But where facts are important, key information is sometimes not present. This problem proves to be more pronounced in the summarization of long context tasks. One of the factors contributing to this is the positional bias in LMs, where more recent tokens play a more critical role in the generation of the next token~\citep{long_context_llm}. It is also established that summaries are more faithful to tokens at the start and the end~\citep{lm_positional_bias}, ignoring the text in the middle. This phenomenon may be more pronounced in small to medium size LMs~\citep{llama3,gemma2,mistral} because they don't have the scale to extract deep contextual information effectively, leading to the generation of sub-optimal summaries. One way to handle this is by adapting LMs to the task via weight updates (fine-tuning), but this poses massive compute and data requirements, and does not generalize to all use cases. One can also do complex pipelining, or iterative refinement~\citep{summit}, but these bring additional complications and inefficiencies.

However, generating fluent, factually correct, contextually rich summaries efficiently is important to support generalization and scalability. This poses the following key challenges: (\textbf{C1}) How to mitigate the positional bias of LM to generate a good summary? (\textbf{C2}) How to make this approach generalize so that it can work with the summarization of all types of text? (\textbf{C3}) How to achieve this with resource efficiency?

To solve these challenges, we propose \textit{QA-Prompting} -- an approach to summarization by using question-answering and in-context learning with a single LM call. First, we show that vanilla prompting or in-context learning (ICL) generates sub-optimal summaries of articles. Then, using a manually crafted, potentially relevant set of questions, we identify top $k$ questions that can aid summarization. Question answering is different from summarization, as it tries to focus on a specific part of the article rather than comprehending its global context at once. We use this property to extract relevant content from the article. Then, using the article and these questions, we prompt the LM to first generate answers to the questions, followed by generating a summary. By trying to answer the relevant questions first, the LM extracts useful information and keeps it in recent context, which further helps it to generate good summaries (addressing C1). It also filters out noise from the long context article that is not relevant for the summary. Using small LMs in the 0.5B-12B range, we show that this approach significantly improves the quality of summaries (addressing C2). QA-prompting uses a single LM call for a task and works with a pre-trained model without modifications (addressing C3). We also conduct a detailed ablation study to validate our design choices. Our questions are domain-specific, i.e., they are different for tasks belonging to different domains (e.g., news and research). Identifying top $k$ questions is an overhead, but needs to be done only once for a domain. We show that this domain-level adoption of QA-prompting is necessary and performs better than keeping a standard set for all tasks. 

The \textbf{key contributions} of this work is:
\begin{itemize} 
    \item \textit{QA-prompting} -- a novel domain-adaptive QA-driven prompting for text summarization to improve the extraction of critical information. 
    \item Leveraging question-answering as an intermediate step to generate summaries that are both contextually rich and factually accurate in a single LM call. 
    \item Demonstrating its effectiveness with pre-trained LMs in 0.5B-12B, achieving $9-30\%$ improvement in ROUGE scores. 
\end{itemize}

\section{Related Work}

\textbf{Summarization using LLMs}: LLMs have produced state-of-the-art results in abstractive summarization and have significantly accelerated the research in this area~\citep{summarization_almost_dead}. Their summaries have been widely accepted through various works~\citep{news_summarization,benchmarking_llm_summarization}. SummIt~\citep{summit} iteratively prompts ChatGPT to generate a summary and keep improving it based on feedback from an evaluator LLM. But, it is inefficient and costly due to multiple LM calls. Chain-of-density~\citep{chain_of_density} gradually improves summaries using GPT-4 by iteratively adding more content while keeping the length constant. This is also inefficient and makes the summary less readable.

\textbf{Positional bias in LLMs}: In transformer models, earlier and recent tokens are known to dominate the prediction of the next token~\citep{lm_positional_bias,long_context_llm}. Streaming LLM~\citep{streamingllm} was able to generate high-quality text, using just an attention sink and a local attention window. Local context also dominates LM performance in multiple choice QA~\citep{pos_mcq1,pos_mcq2} and arithmetic tasks~\citep{positional_arithmetic}.

\textbf{Prompting and in-context learning}: Prompting emerged as an effective way of utilizing LMs without fine-tuning~\citep{gpt2}. GPT-3~\citep{gpt3} introduced in-context learning (ICL), showing that giving LMs some pre-engineered examples helps them to understand the task better and generate more representative text. A survey related to prompt engineering~\citep{prompt_survey} details extensively on how to prompt LMs for different tasks. Recent works also showed that ICL leads to better out-of-domain (OOD) generalization~\citep{reliable_prompting}. COT prompting~\citep{cot_prompting} showed that performance on reasoning tasks can be improved if a rationale is generated before the answer. ICL remains a dominant strategy since it doesn't require any weight updates and allows using the same model for different tasks.

Our work is motivated by the intersection of these ideas.

\begin{table*}[htbp]
    \small
    \centering
        \begin{tabularx}{\textwidth}{>{\raggedright\arraybackslash}p{2.5cm}X}
            \toprule
            \textbf{Key} & \textbf{Question} \\
            \midrule
            \textbf{topic} & What is the main topic or focus of the content? \\
            \textbf{key\_pts} & What are the key points or arguments presented? \\
            \textbf{entities} & Who are the 3 main entities or individuals involved, and what roles do they play? \\
            \textbf{timeline} & Which timeline, if any, is being discussed here? \\
            \textbf{details} & What are the supporting details, examples, or evidence provided? \\
            \textbf{conclude} & What conclusions, impacts, or implications are mentioned, if any? \\
            \textbf{tone} & What is the overall tone or sentiment (e.g., objective, critical, positive, etc.)? \\
            \textbf{challenges} & What questions or challenges does the content raise? \\
            \textbf{insights} & What unique insights or perspectives are offered? \\
            \textbf{audience} & What audience is the content aimed at, and how does this affect its presentation? \\
            \bottomrule
        \end{tabularx}
    \caption{Candidate questions that are considered for QA-prompting.}
    \label{tab:questions}
\end{table*}

\section{Method Overview}

In this section, we will describe QA-prompting in detail. The first step is to sample candidate questions that will aid the generation of effective summaries, followed by using these questions to construct a prompt that will summarize the text.

\subsection{Sampling Candidate Questions}

The first step is to find relevant questions which will aid the generation of effective summaries. For this purpose, we start with a set of 10 manually crafted questions that we feel might be relevant in all domains, which are listed in Table~\ref{tab:questions}.

Thereafter, for all questions $q_i$, we prompt an LM with the article and $q_i$, asking it to generate the answer $a_i$. We then use it to find its overlap precision $P_i(r, a_i)$ (equation~\ref{eqn:overlap_precision}), which can be defined as the ratio of the number of intersecting words in the generated answer $a_i$ and the reference summary $r$, to the total number of words generated in $a_i$. The intention is to find questions that are relevant for the generation of a summary closer to the reference. This metric may not give a complete evaluation of question-answering, but we don't need that. We only need to rank all ten candidate questions.

\begin{equation}
    P_i(r, a_i) = \frac{|W(a_i) \cap W(r)|}{|W(a_i)|}
    \label{eqn:overlap_precision}
\end{equation}

\begin{equation}
    W(x) = \text{ number of words in } x
\end{equation}

From this step, for each LM and domain pair, we find the most to least important questions as per the decreasing order of overlap precision. In the next step, we will show how we use this result to construct our prompt for summarization.

\subsection{QA-prompting}

After we have ranked the order of importance of questions for each domain, we select top $k$ questions for our summarization prompt, $k$ being a user-defined hyperparameter. With the set of $k$ questions,  we prompt the LM to first answer the questions and then generate the summary. The exact prompt is detailed below.

\begin{mdframed}
\texttt{Given the following article, first answer the questions. Then, using the article and answers as key pointers, generate a summary of the article. \\ \{article\} \\ Q1: \{question\_1\} \\ Q2: \{question\_2\} \\ ... \\ Qk: \{question\_k\} \\ A:}
\end{mdframed}

The questions are arranged from maximum overlap precision score to the minimum selected. It may seem that adding the highest overall precision question should be added at the end, given the positional bias of transformers. But, through experiments, we empirically found that this order performs slightly worse. It may be because generating the answers to more important questions first also aids the LM in generating better answers to subsequent questions, thereby contributing to better summaries overall.

To guide the model on how to proceed with the generation of the answers, we provide in-context examples with completed answers and summaries. The examples are taken from the same task to resemble similarity; answers are taken from corresponding generation answers, and the reference summary is included. One example of a structure of output is shown below.

\begin{mdframed}
\texttt{A: A1: \{answer\_1\}. A2: \{answer\_2\}. ... Ak: \{answer\_k\}. \\ Summary: \{summary\}.}
\end{mdframed}

The complete prompt first contains in-context examples followed by the task instance of interest. This is passed to the LM, and the generated text is retrieved. From that, we extract the summary.

\section{Experimental Setup}

This section is to describe the experimental setting to validate our proposed method. All the artifacts used are cited as per their licensing agreements for academic research.

\subsection{Dataset}

To create our experimental dataset, we construct a test set using popular summarization datasets like CNN Dailymail~\citep{cnn_dailymail}, Samsum~\citep{samsum}, Multinews~\citep{multinews}, XSum~\citep{xsum}, PubMed~\citep{pubmed} and other summarization task instances of the Supernatural Instructions~\citep{super_natural_instructions} dataset, which contains Amazon food reviews, dialogue summarization, along with labeled application domains. All of these are widely-used benchmarks released after careful checks of PII or offensive content. The distribution of the experimental data along with domain names is given in Table~\ref{tab:domain_distribution}. The domain classifications were taken directly from Supernatural Instructions. Each domain can have multiple datasets, like CNN/Dailymail, XSum, and all news datasets will be under News.

\begin{table}[htbp]
    \centering
    \begin{tabular}{c|c}
    \toprule
      Domain & Instances  \\
    \midrule
      Commonsense   & 600 \\
      Dialogue & 1200 \\
      News & 3000 \\
      Public Places & 600 \\
      Reviews & 1200 \\
      Research & 600 \\
    \bottomrule
    \end{tabular}
    \caption{Distribution of number of task instances in each domain in the experimental data.}
    \label{tab:domain_distribution}
\end{table}

\subsection{Models}
\label{sec:exp-models}

We experiment with multiple LMs, which include Llama-3.2-1B~\citep{llama3}, Llama-3.1-8B~\citep{llama3}, Mistral-7B~\citep{mistral}, Qwen2.5 family of models~\citep{qwen2.5,qwen2}, and Gemma-3 family of models~\citep{gemma_3}. We use the pre-trained version of each of the models to see how they perform with QA-prompting without any instruction tuning. The intention behind selecting models is to find the patterns in performance with respect to varying differences. For execution, we use a batch size of 4 (8 for 1B), max tokens as $512 + 32 * k$, and use greedy decoding. All models run on a single NVIDIA H200 GPU.

For different types of experiments, we use different subsets of models that fit the settings, which will be detailed in respective subsections.

\subsection{Evaluation and Analysis}
\label{sec:exp-eval}

We evaluate our method with four metrics -- ROUGE-1, ROUGE-2, ROUGE-L~\citep{lin2004rouge}, and BERTScore F1~\citep{zhang2019bertscore} (with Roberta large). BERTScore analysis is important as LM outputs sometimes don't match at the n-gram level but are still semantically correct. ROUGE has known limitations, and some works~\citep{g_eval,sinha-etal-2025-guiding} have emphasized using LLM as a judge to evaluate. But, it is still a robust metric for summarization; with lots of other works using it for reporting results of summarization. We analyze the results on values of $k$ from 0 to 5, and also compare them against baseline prompting for summarization, vanilla in-context learning, and other state-of-the-art (SOTA) techniques.

\section{Results}

This section discusses the results, followed by an ablation study, ending with aspects related to domain-specificity and question selection.

\subsection{QA-prompting}

\begin{table*}[!ht]
\centering
\small
\setlength{\tabcolsep}{5pt}
\begin{tabular}{lllccccc}
\toprule
\textbf{Method} & \textbf{Model Name} & \textbf{Params.} & \textbf{Best k} & \textbf{ROUGE-1} & \textbf{ROUGE-2} & \textbf{ROUGE-L} & \textbf{BERTScore F1} \\
\midrule
\multirow{6}{*}{{Prompting}} & Llama-3.2 & 1B & - & 18.12 & 7.26 & 14.30 & 83.86 \\
                      & Llama-3.1 & 8B & - & 22.93 & 10.91 & 18.03 & 83.19 \\
                      & Mistral & 7B & - & 4.73 & 3.07 & 4.32 & 20.94 \\
                      & Qwen2.5 & 0.5B & - & 22.10 & 9.87 & 17.30 & 85.22 \\
                      & Qwen2.5 & 7B & - & 25.52 & 10.47 & 19.06 & 86.27 \\
                      & Gemma-3 & 12B & - & 8.18 & 3.18 & 6.44 & 40.14 \\
\midrule
\multirow{6}{*}{{ICL}} & Llama-3.2 & 1B & - & 27.87 & 14.54 & 24.20 & 71.86 \\
                      & Llama-3.1 & 8B & - & 30.07 & 16.78 & 26.32 & 83.20 \\
                       & Mistral & 7B & - & 34.86 & 19.82 & 30.67 & 83.79 \\
                      & Qwen2.5 & 0.5B & - & 26.75 & 11.53 & 21.19 & 86.04 \\
                      & Qwen2.5 & 7B & - & 28.46 & 12.63 & 21.90 & 85.86 \\
                      & Gemma-3 & 12B & - & 33.11 & 18.48 & 32.91 & 83.37 \\
\midrule
       \citet{summit} & ChatGPT & 175B & - & 37.29 & 13. 60 & 26.87 & N/A \\
\midrule
       \multirow{3}{*}{{\citet{element-aware-summ}}} & BART-Large & 406M & - & 30.89 & 11.59 & 26.12 & 87.85 \\
       & T5-Large & 770M & - & 31.23 & 12.28 & 27.15 & 87.48 \\
       & GPT-3.5 & 175B & - & 34.75 & 13.08 & 29.84 & 89.19 \\
\midrule
       \multirow{3}{*}{{\citet{hadas}}} & Flan-T5-S & 80M & - &  N/A & N/A & 17.16 & N/A \\
       & Flan-T5-B & 250M & - &  N/A & N/A & 18.77 & N/A \\
       & BART-base & 139M & - &  N/A & N/A & 23.62 & N/A \\
\midrule
       \multirow{3}{*}{{\citet{rl_summarization}}} & GPT-J & 6B & - &  N/A & N/A & 25.68 & N/A \\
       & Mistral & 7B & - &  N/A & N/A & 27.98 & N/A \\
       & Llama-2 & 7B & - &  N/A & N/A & 27.24 & N/A \\
\midrule
       \multirow{3}{*}{{\citet{xu-etal-2024-salient}}} & Claude & x & - & 42.78 & N/A & 28.23 & N/A \\
       & Mistral & 7B & - & \textbf{43.45} & N/A & 27.83 & N/A \\
       & Falcon & 40B & - & 36.70 & N/A & 25.85 & N/A \\
\midrule
\multirow{6}{*}{\textbf{Ours (ICL+QA)}} & Llama-3.2 & 1B & 2 & 31.14 & 15.49 & 27.35 & 78.02 \\
                      & Llama-3.1 & 8B & 2 & 40.51 & 21.06 & 34.14 & 89.17 \\
                      & Mistral & 7B & 2 & 41.97 & \textbf{21.82} & 35.92 & 90.09 \\
                      & Qwen2.5 & 0.5B & 1 & 28.15 & 14.46 & 23.07 & 86.21 \\
                      & Qwen2.5 & 7B & 3 & 31.66 & 15.54 & 26.92 & 80.43 \\
                      & Gemma-3 & 12B & 4 & {43.12} & 21.49 & \textbf{38.92} & \textbf{90.44} \\
\bottomrule
\end{tabular}
\caption{Mean ROUGE Scores (0-100) and BERTScore F1 (0-100) for various models averaged over entire experimental set of all methods. QA-prompting consistently outperforms vanilla prompting (Prompting), in-context learning (ICL), and other methods (N/A = Not Available, x=Unknown).}
\label{tab:results}
\end{table*}

First, we report the performance of all models discussed in section~\ref{sec:exp-models} and compare it against baselines as described in section~\ref{sec:exp-eval}. The results are tabulated in table~\ref{tab:results}. The prompting, ICL, and QA-prompting results were calculated by us, and we took the results of other papers directly as reported.

We see that our method consistently outperforms vanilla prompting and ICL. Some interesting patterns observed are that Mistral-7B and Gemma-3-12B perform unacceptably bad using vanilla prompting. But, once they get in-context examples, the performance is much better. After QA-prompting, the result further improves. 

For small models ($\leq 1B$) like Llama-3.2-1B and Qwen-2.5-0.5B, the ROUGE-L gain from ICL to QA-prompting is the least. For Llama-3.2-1B, performance almost doubles from vanilla prompting to baseline in-context learning (ICL), and the ROUGE-L gain from ICL to our method is of only $13.02\%$. Similarly, for Qwen2.5-0.5B, it is $8.87\%$. We believe this behavior occurs because these models can't use the extracted information to improve summaries due to limitations of scale.

This pattern is also visible in other models; however, the extent of increase between vanilla prompting and ICL decreases as the model size increases. This is with the exception of Mistral-7B and Gemma-3-12B, which seem to not understand the task properly and drastically underperform when using vanilla prompting. Simultaneously, the increase between ICL to QA-prompting remains high, with $18.29\%$ gain on Gemma-3-12B and $29.75\%$ gain on the Llama-3.1-8B model. Mistral-7B and Qwen-2.5-7B also witness an increase of $17.10\%$ and $22.95\%$ respectively. The optimal $k$ also roughly increases as model complexity increases. We can, therefore, claim with reasonable confidence that increasing model size increases the extent of improvement using QA-prompting. This may be due to increasing model complexity leading to better utilization of information from the answers to generate a better summary. However, the gain differs between the models, which may be coming from their different inherent properties. Some qualitative results are given in the appendix~\ref{sec:appendix}.

\subsubsection{Comparison with State-of-the-art}

We compare QA-prompting against baselines and other state-of-the-art (SOTA) methods, which are tabulated in Table~\ref{tab:results}. For other methods, all values are averaged over all datasets that the individual works report results on. Also, their models and datasets are different from ours. We use a broader benchmark and a larger set of models.

We can see that our method consistently outperforms other techniques. Compared to SummIt~\citep{summit}, which iteratively improves summarization, our method uses smaller models, makes only 1 model call, and outperforms it by $~45\%$ (with Gemma 3). The most comparable set is ~\citet{rl_summarization},~\citet{xu-etal-2024-salient} and QA-prompting with Mistral 7B, where QA-prompting performs better by $~28\%$. It also outperforms element-aware summarization~\citep{element-aware-summ} using a large model like GPT-3.

In the following subsections, we will conduct a detailed ablation study of QA-prompting. As we divide the approach domain-wise, we will look at the results of each domain. We will also show that the domain-specific QA-prompting is better compared to using a generic set of questions.

\subsubsection{Performance v/s Model Scale}

\begin{figure}
    \centering
    \includegraphics[width=\linewidth]{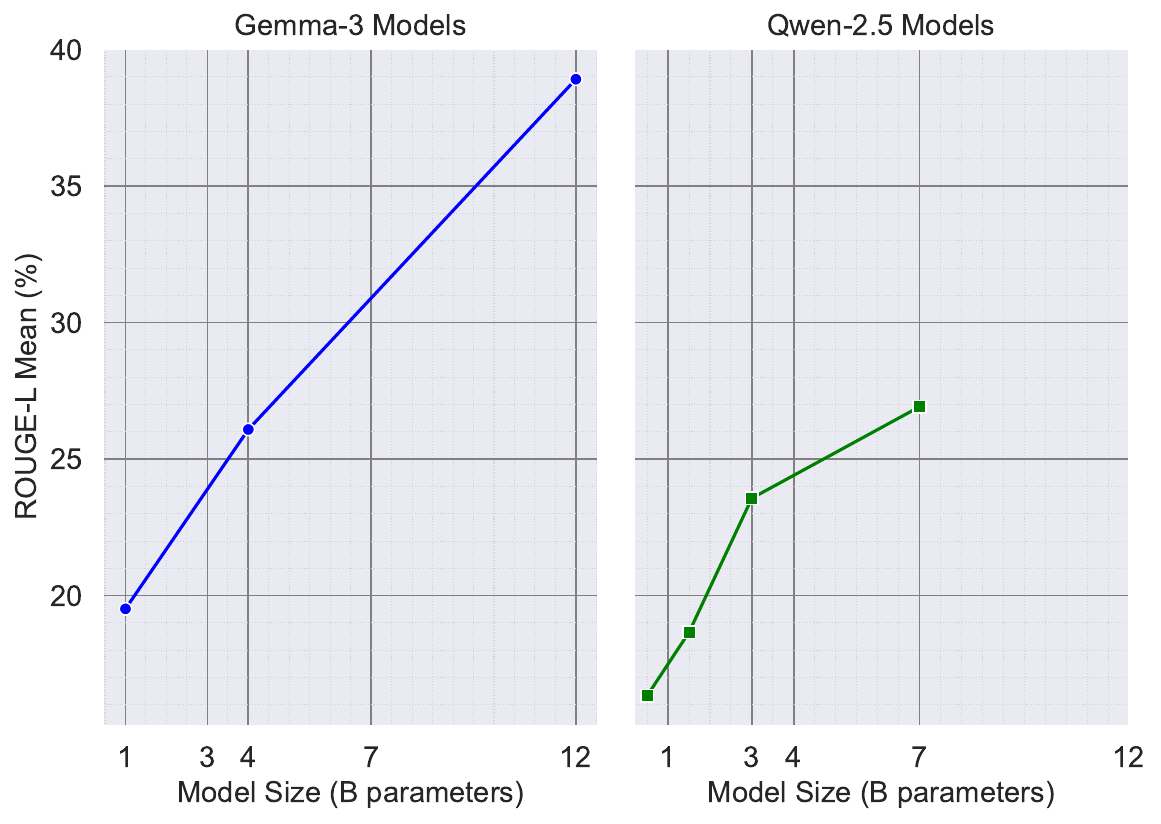}
    \caption{Variation of Rouge-L score with scale of model for $k=3$ for Qwen 2.5 family (best k for 7B), and $k=4$ for Gemma 3 family (best $k$ for Gemma-3-12B). Performance improves with scale.} 
    \label{fig:model_scale_vs_rougeL}
\end{figure}

The above results give an indication that the best $k$ increases as model size increases. However, to concretely understand the variation of performance with model scale, we experiment with Qwen2.5 (0.5B, 1.5B, 3B, 7B) and the Gemma-3 family (1B, 4B, 12B) with fixed $k$ and analyze the variation in performance. We choose these models because they give multiple models at different scales. The results are shown in Figure~\ref{fig:model_scale_vs_rougeL}.

We can see that the performance improves as the scale of parameters increases. This shows that QA-prompting is able to extract useful information from questions to generate the summary. The rate of increase almost remains the same for Gemma-3. For Qwen2.5, there is a steep increase from 0.5B to 3B, and then it is relatively less.

\subsubsection{Performance v/s `$k$'}

\begin{figure}[!ht]
    \centering
    \begin{subfigure}{\linewidth}
        \centering
        \includegraphics[width=\linewidth]{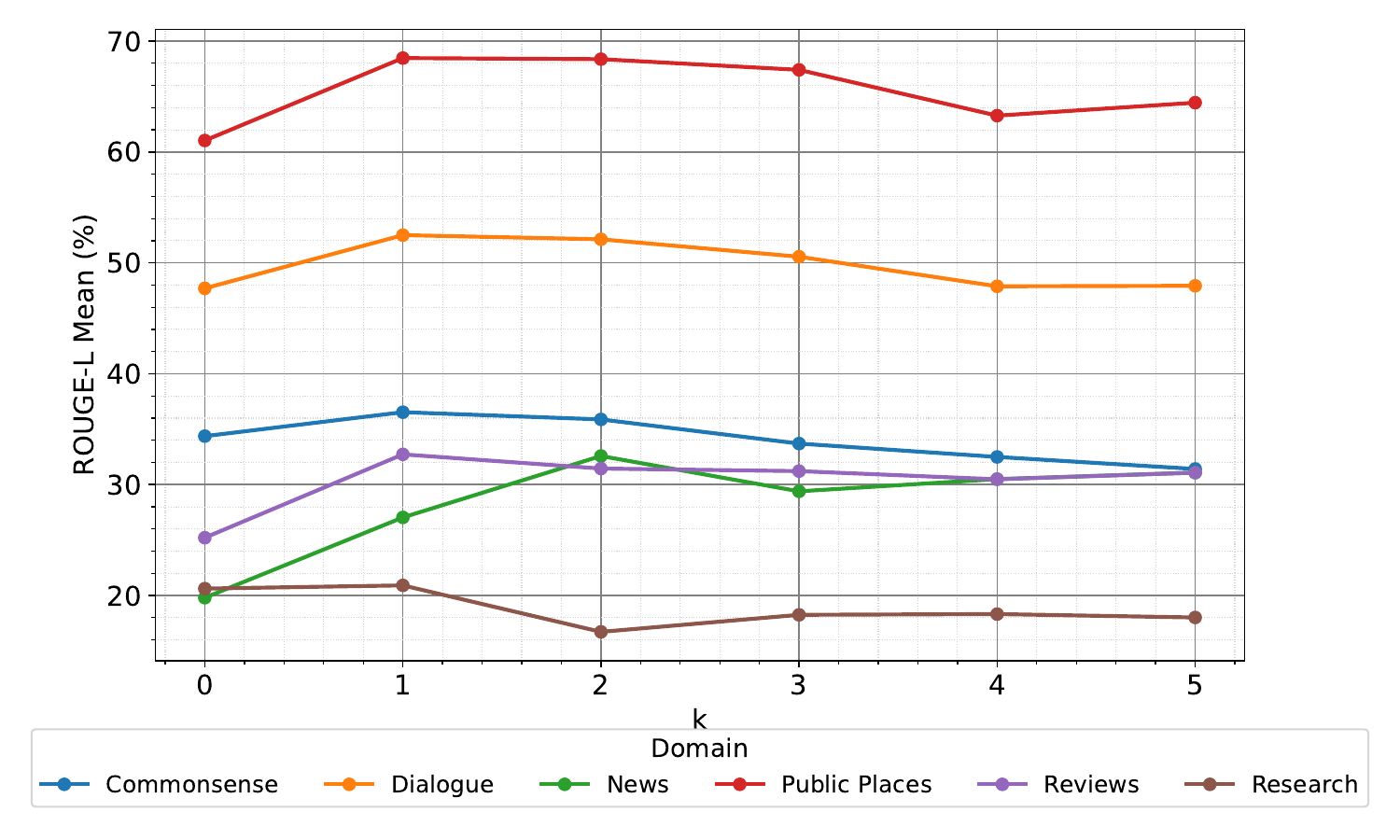}
        \caption{Llama-3.1-8B}
        \label{fig:k_vs_rougeL_llama}
    \end{subfigure}
    \vspace{0.5cm} 
    \begin{subfigure}{\linewidth}
        \centering
        \includegraphics[width=\linewidth]{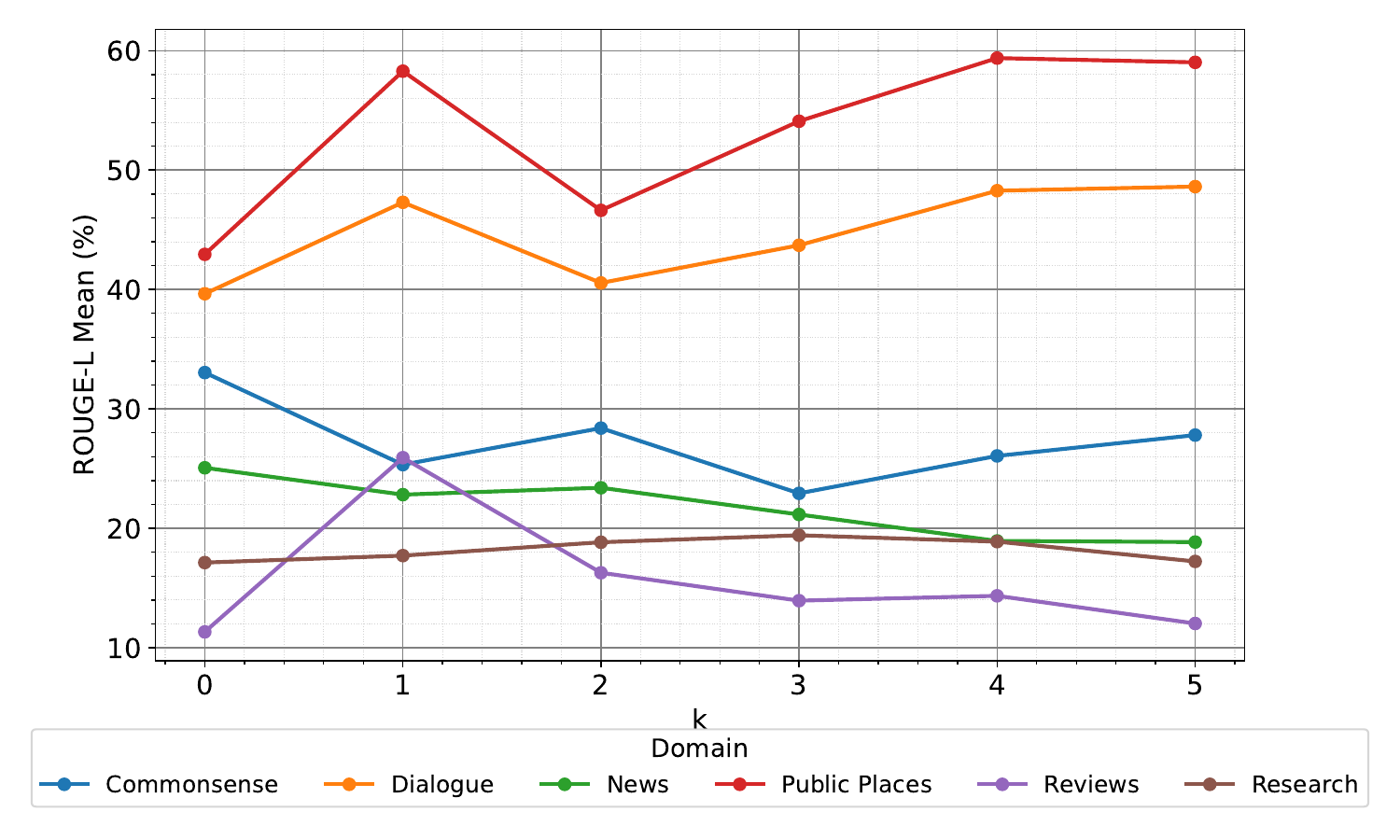}
        \caption{Gemma-3-12B}
        \label{fig:k_vs_rougeL_gemma}
    \end{subfigure}
    \caption{Variation of Rouge-L score with top-$k$ questions across different domains for Llama-3.1-8B and Gemma-3-12B.}
    \label{fig:k_vs_rougeL}
\end{figure}

To analyze the impact of $k$, i.e., the number of questions considered for QA-prompting, we plot the variation of the ROUGE-L score of each domain for $k=0,1,2,3,4,5$ for Llama-3.1-8B and Gemma-3-12B. We select these to show certain variations in behavior which we will discuss below. The results are visualized in figure~\ref{fig:k_vs_rougeL}. 

First, we can infer from the figure that optimal values of $k$ are different for different domains for both (and all other) models. For example, with Llama-3.1-12B, for news articles, $k=5$ performs the best, for research, $k=1$ is the best, and for commonsense articles, $k=2$ is the best. The $k^{th}$ questions will also be different for different domains. This shows that different domains require different questions and numbers of questions for optimal summaries. We also observed that the trend of variation differs for each domain. For example, for news articles, the performance continues to rise till 2, gets a sudden dip at 3, and then again increases.

The tasks where the performance is high at $k=0$ show that the model is inherently better in these tasks without QA prompting. But, at other places, QA prompting is contributing. Different models have different strengths too. For example, Llama is better for reviews, and Gemma-3 struggles more with it. On the other hand, it is the opposite with research articles.

\subsubsection{Performance v/s Context Size}

\begin{figure}
    \centering
    \includegraphics[width=\linewidth]{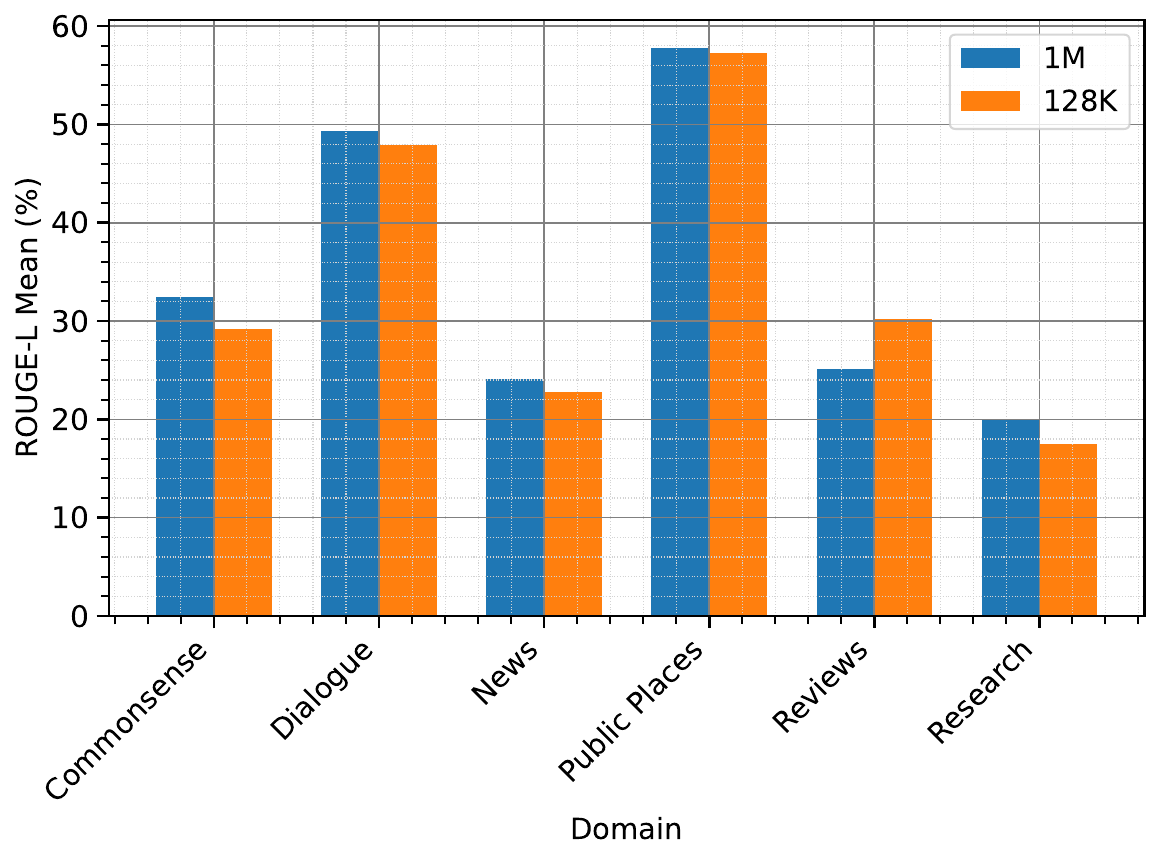}
    \caption{Rouge-L score for Qwen-2.5-7B (instruction-tuned) with 1M and 128K context size for different domains. Performance doesn't decrease significantly}
    \label{fig:context_size_vs_rougeL}
\end{figure}

QA-prompting requires taking the article, generating the answers, and then generating the summary. This can lead to large context size requirements. While models are being introduced with 1M context size, most of them are in the range of 8K-128K. So, it's important to determine how much the increased context size helps QA-prompting, and what the relative limitations of LMs with smaller context are. For this, we experiment with the Qwen2.5-7B model which is available in both $128K$ and $1M$ context sizes and report their Rouge-L scores for all domains in Figure~\ref{fig:context_size_vs_rougeL}.

From the results, we can see that even when the context size is reduced by $87.2\%$, the performance across domains didn't decrease by more than $10\%$. Also, while the performance decreases with decreasing context size for most domains, it improves for reviews. We believe this is because the reviews are short, and a smaller context size model helps it since it is able to focus better on shorter text. 

Intuition may suggest that since we are increasing the generation length by doing question-answering before summarization, a larger context size may be required. But, the empirical results show otherwise. From this, we can also validate the positional behavior of LMs. Useful information in recent context allows the model to generate better summaries even with context size limitations. Therefore, QA-prompting can be used to generate a better summary of long-context tasks using shorter context-sized models.

\subsection{Domain Specificity of QA-prompting}

Some questions that emerge with this approach are: (\textbf{Q1}) are domain-specific questions really required for QA-prompting, or is a general set of questions sufficient? \textbf{(Q2)} Do the questions really differ for different domains? \textbf{(Q3)} Are these differences, if any, consistent across models? This section will try to answer these.

\begin{table}[htbp]
    \centering
    \begin{tabular}{c|c|c|c|c}
    \toprule
        Model & $k$ & ICL & QA-G & QA-DS  \\
    \midrule
        Gemma-3-12B & 4 & 32.91 & 36.76 & 38.92 \\
        Mistral-7B & 2 & 30.67 & 31.44 & 35.92 \\
    \bottomrule
    \end{tabular}
    \caption{Mean ROUGE-L score of in-context Learning (ICL), domain-Specific QA-prompting (QA-DS), and global QA-prompting (QA-G) for Gemma-3-12B and Mistral-7B, using the top-$k$ questions.}
    \label{tab:ds_vs_g_prompting}
\end{table}

To address Q1, in addition to selecting top-$k$ domain-specific questions, we also collect top-$k$ globally best questions and perform QA-prompting using them. This is done for Gemma-3-12B, our best performing model, and Mistral-7B, the second best. We tabulate the results of in-context learning (ICL), domain-specific QA-prompting (QA-DS), and global QA-prompting (QA-G) in Table~\ref{tab:ds_vs_g_prompting}. We had earlier hypothesized that domain-specificity adds performance by allowing models to extract relevant information, which can differ between domains. For example, research articles may find insights and challenges to be more relevant, whereas news articles may find entities involved and key points to be more relevant. Here, we empirically find that to be correct, with $5.88\%$ gain from domain-specificity in Gemma-3-12B, and $14.25\%$ in Mistral-7B. We believe that the difference in increase percentage can come from two factors - higher value of $k$ being optimal for Gemma-3 means that it extracts more information, and Gemma being more expressive can better extract information and suppress noise/confusion better than Mistral. 

\begin{table*}[htbp]
    \footnotesize
    \centering
    \renewcommand{\arraystretch}{0.9}
    \begin{tabular}{l@{\hskip 2.5pt}|c@{\hskip 2pt}|c@{\hskip 2pt}|c@{\hskip 2pt}|c@{\hskip 2pt}|c@{\hskip 2pt}|c@{\hskip 2pt}|c@{\hskip 2pt}|c@{\hskip 2pt}|c@{\hskip 2pt}|c@{\hskip 2pt}}
        \toprule
        \textbf{Model} & \textbf{Topic} & \textbf{Key Pts} & \textbf{Entities} & \textbf{Timeline} & \textbf{Details} & \textbf{Conclude} & \textbf{Tone} & \textbf{Challenges} & \textbf{Insights} & \textbf{Audience} \\
        \midrule
        Llama-3.2-1B & \textcolor{ForestGreen}{1} & \textcolor{ForestGreen}{3} & \textcolor{ForestGreen}{2} & \textcolor{red}{8} & \textcolor{red}{6} & \textcolor{ForestGreen}{5} & \textcolor{red}{10} & \textcolor{red}{9} & \textcolor{ForestGreen}{4} & \textcolor{red}{7} \\
        Llama-3.1-8B & \textcolor{ForestGreen}{1} & \textcolor{ForestGreen}{3} & \textcolor{ForestGreen}{2} & \textcolor{red}{10} & \textcolor{red}{6} & \textcolor{ForestGreen}{5} & \textcolor{red}{9} & \textcolor{red}{8} & \textcolor{ForestGreen}{4} & \textcolor{red}{7} \\
        Mistral-7B & \textcolor{ForestGreen}{5} & \textcolor{ForestGreen}{1} & \textcolor{red}{6} & \textcolor{red}{7} & \textcolor{ForestGreen}{3} & \textcolor{ForestGreen}{2} & \textcolor{red}{10} & \textcolor{red}{8} & \textcolor{ForestGreen}{4} & \textcolor{red}{9} \\
        Qwen2.5-0.5B & \textcolor{ForestGreen}{1} & \textcolor{ForestGreen}{4} & \textcolor{red}{9} & \textcolor{ForestGreen}{3} & \textcolor{red}{6} & \textcolor{ForestGreen}{2} & \textcolor{red}{7} & \textcolor{red}{8} & \textcolor{ForestGreen}{5} & \textcolor{red}{10} \\
        Qwen2.5-7B & \textcolor{ForestGreen}{2} & \textcolor{ForestGreen}{3} & \textcolor{red}{8} & \textcolor{ForestGreen}{4} & \textcolor{ForestGreen}{5} & \textcolor{ForestGreen}{1} & \textcolor{red}{9} & \textcolor{red}{7} & \textcolor{red}{6} & \textcolor{red}{10} \\
        Gemma-3-12B & \textcolor{ForestGreen}{1} & \textcolor{ForestGreen}{2} & \textcolor{ForestGreen}{4} & \textcolor{red}{9} & \textcolor{ForestGreen}{3} & \textcolor{red}{6} & \textcolor{red}{8} & \textcolor{red}{7} & \textcolor{ForestGreen}{5} & \textcolor{red}{10} \\
        \bottomrule
    \end{tabular}
    \caption{Ranking of various questions for different models. Numbers in \textcolor{ForestGreen}{GREEN ($\leq5$)} may be considered in our experiments (since we experiment till k=5); numbers in \textcolor{red}{RED ($>5$)} are ignored.}
    \label{tab:question_rankings_model}
\end{table*}

\begin{table*}[htbp]
    \footnotesize
    \centering
    \renewcommand{\arraystretch}{0.9}
    \begin{tabular}{l@{\hskip 2.5pt}|c@{\hskip 2pt}|c@{\hskip 2pt}|c@{\hskip 2pt}|c@{\hskip 2pt}|c@{\hskip 2pt}|c@{\hskip 2pt}|c@{\hskip 2pt}|c@{\hskip 2pt}|c@{\hskip 2pt}|c@{\hskip 2pt}}
        \toprule
        \textbf{Domain} & \textbf{Topic} & \textbf{Key Pts} & \textbf{Entities} & \textbf{Timeline} & \textbf{Details} & \textbf{Conclude} & \textbf{Tone} & \textbf{Challenges} & \textbf{Insights} & \textbf{Audience} \\
        \midrule
        Commonsense & \textcolor{ForestGreen}{3} & \textcolor{ForestGreen}{2} & \textcolor{ForestGreen}{5} & \textcolor{ForestGreen}{4} & \textcolor{red}{6} & \textcolor{ForestGreen}{1} & \textcolor{red}{8} & \textcolor{red}{7} & \textcolor{red}{9} & \textcolor{red}{10} \\
        Dialogue & \textcolor{ForestGreen}{4} & \textcolor{ForestGreen}{5} & \textcolor{red}{7} & \textcolor{ForestGreen}{2} & \textcolor{ForestGreen}{6} & \textcolor{red}{8} & \textcolor{ForestGreen}{1} & \textcolor{ForestGreen}{3} & \textcolor{red}{9} & \textcolor{red}{10} \\
        News & \textcolor{ForestGreen}{2} & \textcolor{ForestGreen}{1} & \textcolor{ForestGreen}{3} & \textcolor{ForestGreen}{4} & \textcolor{ForestGreen}{5} & \textcolor{red}{6} & \textcolor{red}{9} & \textcolor{red}{8} & \textcolor{red}{7} & \textcolor{red}{10} \\
        Public Places & \textcolor{ForestGreen}{2} & \textcolor{ForestGreen}{4} & \textcolor{red}{7} & \textcolor{red}{10} & \textcolor{ForestGreen}{5} & \textcolor{ForestGreen}{6} & \textcolor{red}{9} & \textcolor{ForestGreen}{3} & \textcolor{red}{8} & \textcolor{ForestGreen}{1} \\
        Reviews & \textcolor{ForestGreen}{4} & \textcolor{red}{7} & \textcolor{ForestGreen}{2} & \textcolor{ForestGreen}{5} & \textcolor{red}{8} & \textcolor{ForestGreen}{6} & \textcolor{ForestGreen}{1} & \textcolor{ForestGreen}{3} & \textcolor{red}{9} & \textcolor{ForestGreen}{2} \\
        Research & \textcolor{ForestGreen}{3} & \textcolor{ForestGreen}{2} & \textcolor{red}{7} & \textcolor{ForestGreen}{5} & \textcolor{ForestGreen}{6} & \textcolor{red}{8} & \textcolor{ForestGreen}{4} & \textcolor{red}{9} & \textcolor{red}{10} & \textcolor{ForestGreen}{1} \\
        \bottomrule
    \end{tabular}
    \caption{Ranking of various questions for different domains for Mistral-7B. Numbers in \textcolor{ForestGreen}{GREEN ($\leq5$)} may be considered in our experiments (since we experiment till k=5); numbers in \textcolor{red}{RED ($>5$)} are ignored.}
    \label{tab:question_rankings_domain}
\end{table*}

To address Q3, we rank the questions for each of the models by individually generating their answers, and then calculating the overlap precision score using the reference summary, as defined in equation~\ref{eqn:overlap_precision}. Note, we don't need quantitative values of this, as discussed earlier. We are using this metric coarsely to rank the relevance of different questions. So, it is better to analyze the results as ranks. We report these ranks averaged over all domains for different models in Table~\ref{tab:question_rankings_model}.

From the results, we can see that the variation of importance of questions varies significantly with different models. For example, Gemma-3-12B and Mistral-7B are our best models, but the rank of the `topic' question is fifth and first respectively. Since we experiment with a maximum of top 5 questions, some questions like tone, challenges, and audience are never used. They are consistently in the bottom three. Multiple other patterns can be found, showing high variance of the rank of questions for different models, answering \textbf{Q3}.

Similar to above, to answer \textbf{Q2}, we rank the questions for different domains for Mistral-7B. We choose Mistral-7B because it is our second best-performing model as per Table~\ref{tab:results}, has lot of other benchmarks to compare, and also shows a significant gain of $14.25\%$ when using domain specificity, as per Table~\ref{tab:ds_vs_g_prompting}. The results clearly indicate that different domains require different questions to extract better summaries. The differences are probably more than model-level variation. For example, `audience' question varies from rank 10 (worst) to rank 1 (best). This can conclude that questions really differ across models.

\section{Conclusion}

In this work, we present QA-prompting, an effective approach to improve summarization with large language models by incorporating question-answering as an intermediate step. Our method addresses positional bias in language models by extracting key information through domain-specific questions, ensuring critical details remain in recent context before summary generation. Experiments across multiple models and diverse domains demonstrate that QA-prompting consistently outperforms other methods by up to $29\%$.

QA-prompting’s success lies in domain-specific question selection. Tailored question sets significantly enhance performance, especially for larger models that can better process and utilize extracted information. Our efficient method uses pre-trained models without instruction-tuning or fine-tuning, operating in a single LM call and making it scalable for real-world applications.

Future research directions include automating question selection and exploring dynamic top-$k$ optimization for different tasks. By bridging the gap between question-answering and summarization, our work opens new possibilities for leveraging intermediate reasoning steps to improve LM performance across diverse applications.

\section*{Limitations}

While QA-prompting demonstrates significant improvements in summarization quality, there are some limitations that warrant discussion.

\textbf{Domain-Specific Question Selection}: The effectiveness of QA-prompting relies heavily on the relevance of the selected questions to the target domain. While we show that domain-specific questions improve performance, manually curating these questions for new domains and ranking them requires human effort and expertise. Automated methods for question generation or selection could help address this limitation.

\textbf{Model Scale Dependency}: Our experiments reveal that the benefits of QA-prompting increase with model scale. Smaller models show limited gains, suggesting that the approach may be less effective for resource-constrained applications that require very small models.

\textbf{Question-Answering Quality}: The quality of the intermediate question-answering step directly impacts summary quality. Errors or hallucinations in the generated answers could propagate to the final summary. While we mitigate this through question selection, the approach remains vulnerable to LM inaccuracies.

\textbf{Single-Pass Generation}: QA-prompting performs question-answering and summarization in a single forward pass. While efficient, this may limit the depth of information extraction compared to multi-step approaches that could refine answers iteratively or via pipelines.

These limitations suggest directions for future work, including automated question generation, hybrid approaches combining QA-prompting with iterative refinement, and better evaluation methodologies. Despite these limitations, QA-prompting provides a simple yet effective approach to improving summarization quality across diverse domains.

\section*{Ethical Considerations}

This work relies on publicly available datasets and pre-trained language models, ensuring no new data collection or human annotation was required. While the datasets used are widely adopted in NLP research, we acknowledge that they may contain biases or sensitive content inherent to their sources. However, as our method operates on existing benchmarks without modification, we did not perform additional bias mitigation or content filtering. 

The proposed QA-prompting approach is designed for abstractive summarization and should not be deployed in high-stakes domains (e.g., legal or medical) without further validation in the use case of interest, as errors in question-answering could propagate to summaries. All experiments were conducted using standard evaluation protocols, and model outputs were analyzed only for research purposes. 

\section*{Acknowledgment}

This work is done as part of Master’s project in the School of Interactive Computing, College of Computing at the Georgia Institute of Technology (Georgia Tech). I am grateful to my advisor, Prof. Alan Ritter, for his guidance, feedback, and encouragement throughout this project. Computational resources were generously provided by the PACE (Partnership for an Advanced Computing Environment) cluster at Georgia Tech, which enabled efficient experimentation with large language models with their wide range and availabilities of GPUs.

\bibliography{custom}

\begin{thebibliography}{41}
\providecommand{\natexlab}[1]{#1}

\bibitem[{Adams et~al.(2023)Adams, Fabbri, Ladhak, Lehman, and Elhadad}]{chain_of_density}
Griffin Adams, Alex Fabbri, Faisal Ladhak, Eric Lehman, and No{\'e}mie Elhadad. 2023.
\newblock \href {https://doi.org/10.18653/v1/2023.newsum-1.7} {From sparse to dense: {GPT}-4 summarization with chain of density prompting}.
\newblock In \emph{Proceedings of the 4th New Frontiers in Summarization Workshop}, pages 68--74, Singapore. Association for Computational Linguistics.

\bibitem[{AI@Meta(2024)}]{llama3}
AI@Meta. 2024.
\newblock \href {https://github.com/meta-llama/llama3/blob/main/MODEL_CARD.md} {Llama 3 model card}.

\bibitem[{Brown et~al.(2020)Brown, Mann, Ryder, Subbiah, Kaplan, Dhariwal, Neelakantan, Shyam, Sastry, Askell, Agarwal, Herbert-Voss, Krueger, Henighan, Child, Ramesh, Ziegler, Wu, Winter, Hesse, Chen, Sigler, Litwin, Gray, Chess, Clark, Berner, McCandlish, Radford, Sutskever, and Amodei}]{gpt3}
Tom~B. Brown, Benjamin Mann, Nick Ryder, Melanie Subbiah, Jared Kaplan, Prafulla Dhariwal, Arvind Neelakantan, Pranav Shyam, Girish Sastry, Amanda Askell, Sandhini Agarwal, Ariel Herbert-Voss, Gretchen Krueger, Tom Henighan, Rewon Child, Aditya Ramesh, Daniel~M. Ziegler, Jeffrey Wu, Clemens Winter, and 12 others. 2020.
\newblock Language models are few-shot learners.
\newblock In \emph{Advances in Neural Information Processing Systems}, volume~33, pages 1877--1901.

\bibitem[{Choi et~al.(2024)Choi, Chae, Song, Jo, and Kim}]{rl_summarization}
Jaepill Choi, Kyubyung Chae, Jiwoo Song, Yohan Jo, and Taesup Kim. 2024.
\newblock \href {https://doi.org/10.18653/v1/2024.emnlp-main.1048} {Model-based preference optimization in abstractive summarization without human feedback}.
\newblock In \emph{Proceedings of the 2024 Conference on Empirical Methods in Natural Language Processing}, pages 18837--18851, Miami, Florida, USA. Association for Computational Linguistics.

\bibitem[{Cohan et~al.(2018)Cohan, Dernoncourt, Kim, Bui, Kim, Chang, and Goharian}]{pubmed}
Arman Cohan, Franck Dernoncourt, Doo~Soon Kim, Trung Bui, Seokhwan Kim, Walter Chang, and Nazli Goharian. 2018.
\newblock \href {https://doi.org/10.18653/v1/N18-2097} {A discourse-aware attention model for abstractive summarization of long documents}.
\newblock In \emph{Proceedings of the 2018 Conference of the North American Chapter of the Association for Computational Linguistics: Human Language Technologies, Volume 2 (Short Papers)}, pages 615--621, New Orleans, Louisiana. Association for Computational Linguistics.

\bibitem[{Dong et~al.(2023)Dong, Li, Dai, Zheng, Wu, Chang, Sun, Xu, Li, and Sui}]{icl_survey}
Qingxiu Dong, Lei Li, Damai Dai, Ce~Zheng, Zhiyong Wu, Baobao Chang, Xu~Sun, Jingjing Xu, Lei Li, and Zhifang Sui. 2023.
\newblock \href {https://arxiv.org/abs/2301.00234} {A survey on in-context learning}.
\newblock \emph{Preprint}, arXiv:2301.00234.

\bibitem[{Fabbri et~al.(2019)Fabbri, Li, She, Li, and Radev}]{multinews}
Alexander Fabbri, Irene Li, Tianwei She, Suyi Li, and Dragomir Radev. 2019.
\newblock \href {https://doi.org/10.18653/v1/P19-1102} {Multi-news: A large-scale multi-document summarization dataset and abstractive hierarchical model}.
\newblock In \emph{Proceedings of the 57th Annual Meeting of the Association for Computational Linguistics}, pages 1074--1084, Florence, Italy. Association for Computational Linguistics.

\bibitem[{Gliwa et~al.(2019)Gliwa, Mochol, Biesek, and Wawer}]{samsum}
Bogdan Gliwa, Iwona Mochol, Maciej Biesek, and Aleksander Wawer. 2019.
\newblock \href {https://doi.org/10.18653/v1/D19-5409} {{SAMSum} corpus: A human-annotated dialogue dataset for abstractive summarization}.
\newblock In \emph{Proceedings of the 2nd Workshop on New Frontiers in Summarization}, pages 70--79, Hong Kong, China. Association for Computational Linguistics.

\bibitem[{Jiang et~al.(2023)Jiang, Sablayrolles, Mensch, Bamford, Chaplot, de~las Casas, Bressand, Lengyel, Lample, Saulnier, Lavaud, Lachaux, Stock, Scao, Lavril, Wang, Lacroix, and Sayed}]{mistral}
Albert~Q. Jiang, Alexandre Sablayrolles, Arthur Mensch, Chris Bamford, Devendra~Singh Chaplot, Diego de~las Casas, Florian Bressand, Gianna Lengyel, Guillaume Lample, Lucile Saulnier, Lélio~Renard Lavaud, Marie-Anne Lachaux, Pierre Stock, Teven~Le Scao, Thibaut Lavril, Thomas Wang, Timothée Lacroix, and William~El Sayed. 2023.
\newblock \href {https://arxiv.org/abs/2310.06825} {Mistral 7b}.
\newblock \emph{Preprint}, arXiv:2310.06825.

\bibitem[{Lin(2004)}]{lin2004rouge}
Chin-Yew Lin. 2004.
\newblock Rouge: A package for automatic evaluation of summaries.
\newblock In \emph{Text summarization branches out}, pages 74--81.

\bibitem[{Liu et~al.(2023{\natexlab{a}})Liu, Yuan, Fu, Jiang, Hayashi, and Neubig}]{prompt_survey}
Pengfei Liu, Weizhe Yuan, Jinlan Fu, Zhengbao Jiang, Hiroaki Hayashi, and Graham Neubig. 2023{\natexlab{a}}.
\newblock \href {https://doi.org/10.1145/3560815} {Pre-train, prompt, and predict: A systematic survey of prompting methods in natural language processing}.
\newblock \emph{ACM Comput. Surv.}, 55(9).

\bibitem[{Liu et~al.(2023{\natexlab{b}})Liu, Iter, Xu, Wang, Xu, and Zhu}]{g_eval}
Yang Liu, Dan Iter, Yichong Xu, Shuohang Wang, Ruochen Xu, and Chenguang Zhu. 2023{\natexlab{b}}.
\newblock \href {https://doi.org/10.18653/v1/2023.emnlp-main.153} {{G}-eval: {NLG} evaluation using gpt-4 with better human alignment}.
\newblock In \emph{Proceedings of the 2023 Conference on Empirical Methods in Natural Language Processing}, pages 2511--2522, Singapore. Association for Computational Linguistics.

\bibitem[{Liu et~al.(2022)Liu, Liu, Radev, and Neubig}]{brio}
Yixin Liu, Pengfei Liu, Dragomir Radev, and Graham Neubig. 2022.
\newblock \href {https://doi.org/10.18653/v1/2022.acl-long.207} {{BRIO}: Bringing order to abstractive summarization}.
\newblock In \emph{Proceedings of the 60th Annual Meeting of the Association for Computational Linguistics (Volume 1: Long Papers)}, pages 2890--2903, Dublin, Ireland. Association for Computational Linguistics.

\bibitem[{Narayan et~al.(2018)Narayan, Cohen, and Lapata}]{xsum}
Shashi Narayan, Shay~B. Cohen, and Mirella Lapata. 2018.
\newblock \href {https://doi.org/10.18653/v1/D18-1206} {Don't give me the details, just the summary! topic-aware convolutional neural networks for extreme summarization}.
\newblock In \emph{Proceedings of the 2018 Conference on Empirical Methods in Natural Language Processing}, pages 1797--1807, Brussels, Belgium. Association for Computational Linguistics.

\bibitem[{Ouyang et~al.(2022)Ouyang, Wu, Jiang, Almeida, Wainwright, Mishkin, Zhang, Agarwal, Slama, Gray, Schulman, Hilton, Kelton, Miller, Simens, Askell, Welinder, Christiano, Leike, and Lowe}]{instructGPT}
Long Ouyang, Jeffrey Wu, Xu~Jiang, Diogo Almeida, Carroll Wainwright, Pamela Mishkin, Chong Zhang, Sandhini Agarwal, Katarina Slama, Alex Gray, John Schulman, Jacob Hilton, Fraser Kelton, Luke Miller, Maddie Simens, Amanda Askell, Peter Welinder, Paul Christiano, Jan Leike, and Ryan Lowe. 2022.
\newblock \href {https://openreview.net/forum?id=TG8KACxEON} {Training language models to follow instructions with human feedback}.
\newblock In \emph{Advances in Neural Information Processing Systems}.

\bibitem[{Pezeshkpour and Hruschka(2023)}]{pos_mcq2}
Pouya Pezeshkpour and Estevam Hruschka. 2023.
\newblock \href {https://arxiv.org/abs/2308.11483} {Large language models sensitivity to the order of options in multiple-choice questions}.
\newblock \emph{Preprint}, arXiv:2308.11483.

\bibitem[{Pu et~al.(2023)Pu, Gao, and Wan}]{summarization_almost_dead}
Xiao Pu, Mingqi Gao, and Xiaojun Wan. 2023.
\newblock \href {https://arxiv.org/abs/2309.09558} {Summarization is (almost) dead}.
\newblock \emph{Preprint}, arXiv:2309.09558.

\bibitem[{Radford et~al.(2019)Radford, Wu, Child, Luan, Amodei, Sutskever et~al.}]{gpt2}
Alec Radford, Jeffrey Wu, Rewon Child, David Luan, Dario Amodei, Ilya Sutskever, and 1 others. 2019.
\newblock Language models are unsupervised multitask learners.
\newblock \emph{OpenAI blog}, 1(8):9.

\bibitem[{See et~al.(2017)See, Liu, and Manning}]{cnn_dailymail}
Abigail See, Peter~J. Liu, and Christopher~D. Manning. 2017.
\newblock \href {https://doi.org/10.18653/v1/P17-1099} {Get to the point: Summarization with pointer-generator networks}.
\newblock In \emph{Proceedings of the 55th Annual Meeting of the Association for Computational Linguistics (Volume 1: Long Papers)}, pages 1073--1083, Vancouver, Canada. Association for Computational Linguistics.

\bibitem[{Shen et~al.(2023)Shen, Bubeck, Eldan, Lee, Li, and Zhang}]{positional_arithmetic}
Ruoqi Shen, Sébastien Bubeck, Ronen Eldan, Yin~Tat Lee, Yuanzhi Li, and Yi~Zhang. 2023.
\newblock \href {https://arxiv.org/abs/2311.14737} {Positional description matters for transformers arithmetic}.
\newblock \emph{Preprint}, arXiv:2311.14737.

\bibitem[{Si et~al.(2023)Si, Gan, Yang, Wang, Wang, Boyd-Graber, and Wang}]{reliable_prompting}
Chenglei Si, Zhe Gan, Zhengyuan Yang, Shuohang Wang, Jianfeng Wang, Jordan Boyd-Graber, and Lijuan Wang. 2023.
\newblock \href {https://arxiv.org/abs/2210.09150} {Prompting gpt-3 to be reliable}.
\newblock \emph{Preprint}, arXiv:2210.09150.

\bibitem[{Sinha et~al.(2025{\natexlab{a}})Sinha, Jain, and Chadha}]{sinha-etal-2025-small}
Neelabh Sinha, Vinija Jain, and Aman Chadha. 2025{\natexlab{a}}.
\newblock \href {https://doi.org/10.18653/v1/2025.trustnlp-main.25} {Are small language models ready to compete with large language models for practical applications?}
\newblock In \emph{Proceedings of the 5th Workshop on Trustworthy NLP (TrustNLP 2025)}, pages 365--398, Albuquerque, New Mexico. Association for Computational Linguistics.

\bibitem[{Sinha et~al.(2025{\natexlab{b}})Sinha, Jain, and Chadha}]{sinha-etal-2025-guiding}
Neelabh Sinha, Vinija Jain, and Aman Chadha. 2025{\natexlab{b}}.
\newblock \href {https://aclanthology.org/2025.evalmg-1.7/} {Guiding vision-language model selection for visual question-answering across tasks, domains, and knowledge types}.
\newblock In \emph{Proceedings of the First Workshop of Evaluation of Multi-Modal Generation}, pages 76--94, Abu Dhabi, UAE. Association for Computational Linguistics.

\bibitem[{Sun et~al.(2021)Sun, Krishna, Mattarella-Micke, and Iyyer}]{long_context_llm}
Simeng Sun, Kalpesh Krishna, Andrew Mattarella-Micke, and Mohit Iyyer. 2021.
\newblock \href {https://doi.org/10.18653/v1/2021.emnlp-main.62} {Do long-range language models actually use long-range context?}
\newblock In \emph{Proceedings of the 2021 Conference on Empirical Methods in Natural Language Processing}, pages 807--822, Online and Punta Cana, Dominican Republic. Association for Computational Linguistics.

\bibitem[{Tanya~Goyal(2022)}]{news_summarization}
Greg~Durrett Tanya~Goyal, Junyi Jessy~Li. 2022.
\newblock News summarization and evaluation in the era of gpt-3.
\newblock \emph{arXiv preprint}.

\bibitem[{Team et~al.(2025)Team, Kamath, Ferret, Pathak, Vieillard, Merhej, Perrin, Matejovicova, Ramé, Rivière, Rouillard, Mesnard, Cideron, bastien Grill, Ramos, Yvinec, Casbon, Pot, Penchev, Liu, Visin, Kenealy, Beyer, Zhai, Tsitsulin, Busa-Fekete, Feng, Sachdeva, Coleman, Gao, Mustafa, Barr, Parisotto, Tian, Eyal, Cherry, Peter, Sinopalnikov, Bhupatiraju, Agarwal, Kazemi, Malkin, Kumar, Vilar, Brusilovsky, Luo, Steiner, Friesen, Sharma, Sharma, Gilady, Goedeckemeyer, Saade, Feng, Kolesnikov, Bendebury, Abdagic, Vadi, György, Pinto, Das, Bapna, Miech, Yang, Paterson, Shenoy, Chakrabarti, Piot, Wu, Shahriari, Petrini, Chen, Lan, Choquette-Choo, Carey, Brick, Deutsch, Eisenbud, Cattle, Cheng, Paparas, Sreepathihalli, Reid, Tran, Zelle, Noland, Huizenga, Kharitonov, Liu, Amirkhanyan, Cameron, Hashemi, Klimczak-Plucińska, Singh, Mehta, Lehri, Hazimeh, Ballantyne, Szpektor, Nardini, Pouget-Abadie, Chan, Stanton, Wieting, Lai, Orbay, Fernandez, Newlan, yeong Ji, Singh, Black, Yu, Hui, Vodrahalli, Greff, Qiu,
  Valentine, Coelho, Ritter, Hoffman, Watson, Chaturvedi, Moynihan, Ma, Babar, Noy, Byrd, Roy, Momchev, Chauhan, Sachdeva, Bunyan, Botarda, Caron, Rubenstein, Culliton, Schmid, Sessa, Xu, Stanczyk, Tafti, Shivanna, Wu, Pan, Rokni, Willoughby, Vallu, Mullins, Jerome, Smoot, Girgin, Iqbal, Reddy, Sheth, Põder, Bhatnagar, Panyam, Eiger, Zhang, Liu, Yacovone, Liechty, Kalra, Evci, Misra, Roseberry, Feinberg, Kolesnikov, Han, Kwon, Chen, Chow, Zhu, Wei, Egyed, Cotruta, Giang, Kirk, Rao, Black, Babar, Lo, Moreira, Martins, Sanseviero, Gonzalez, Gleicher, Warkentin, Mirrokni, Senter, Collins, Barral, Ghahramani, Hadsell, Matias, Sculley, Petrov, Fiedel, Shazeer, Vinyals, Dean, Hassabis, Kavukcuoglu, Farabet, Buchatskaya, Alayrac, Anil, Dmitry, Lepikhin, Borgeaud, Bachem, Joulin, Andreev, Hardin, Dadashi, and Hussenot}]{gemma_3}
Gemma Team, Aishwarya Kamath, Johan Ferret, Shreya Pathak, Nino Vieillard, Ramona Merhej, Sarah Perrin, Tatiana Matejovicova, Alexandre Ramé, Morgane Rivière, Louis Rouillard, Thomas Mesnard, Geoffrey Cideron, Jean bastien Grill, Sabela Ramos, Edouard Yvinec, Michelle Casbon, Etienne Pot, Ivo Penchev, and 197 others. 2025.
\newblock \href {https://arxiv.org/abs/2503.19786} {Gemma 3 technical report}.
\newblock \emph{Preprint}, arXiv:2503.19786.

\bibitem[{Team et~al.(2024)Team, Riviere, Pathak, Sessa, Hardin, Bhupatiraju, Hussenot, Mesnard, Shahriari, Ramé, Ferret, Liu, Tafti, Friesen, Casbon, Ramos, Kumar, Lan, Jerome, Tsitsulin, Vieillard, Stanczyk, Girgin, Momchev, Hoffman, Thakoor, Grill, Neyshabur, Bachem, Walton, Severyn, Parrish, Ahmad, Hutchison, Abdagic, Carl, Shen, Brock, Coenen, Laforge, Paterson, Bastian, Piot, Wu, Royal, Chen, Kumar, Perry, Welty, Choquette-Choo, Sinopalnikov, Weinberger, Vijaykumar, Rogozińska, Herbison, Bandy, Wang, Noland, Moreira, Senter, Eltyshev, Visin, Rasskin, Wei, Cameron, Martins, Hashemi, Klimczak-Plucińska, Batra, Dhand, Nardini, Mein, Zhou, Svensson, Stanway, Chan, Zhou, Carrasqueira, Iljazi, Becker, Fernandez, van Amersfoort, Gordon, Lipschultz, Newlan, yeong Ji, Mohamed, Badola, Black, Millican, McDonell, Nguyen, Sodhia, Greene, Sjoesund, Usui, Sifre, Heuermann, Lago, McNealus, Soares, Kilpatrick, Dixon, Martins, Reid, Singh, Iverson, Görner, Velloso, Wirth, Davidow, Miller, Rahtz, Watson, Risdal,
  Kazemi, Moynihan, Zhang, Kahng, Park, Rahman, Khatwani, Dao, Bardoliwalla, Devanathan, Dumai, Chauhan, Wahltinez, Botarda, Barnes, Barham, Michel, Jin, Georgiev, Culliton, Kuppala, Comanescu, Merhej, Jana, Rokni, Agarwal, Mullins, Saadat, Carthy, Perrin, Arnold, Krause, Dai, Garg, Sheth, Ronstrom, Chan, Jordan, Yu, Eccles, Hennigan, Kocisky, Doshi, Jain, Yadav, Meshram, Dharmadhikari, Barkley, Wei, Ye, Han, Kwon, Xu, Shen, Gong, Wei, Cotruta, Kirk, Rao, Giang, Peran, Warkentin, Collins, Barral, Ghahramani, Hadsell, Sculley, Banks, Dragan, Petrov, Vinyals, Dean, Hassabis, Kavukcuoglu, Farabet, Buchatskaya, Borgeaud, Fiedel, Joulin, Kenealy, Dadashi, and Andreev}]{gemma2}
Gemma Team, Morgane Riviere, Shreya Pathak, Pier~Giuseppe Sessa, Cassidy Hardin, Surya Bhupatiraju, Léonard Hussenot, Thomas Mesnard, Bobak Shahriari, Alexandre Ramé, Johan Ferret, Peter Liu, Pouya Tafti, Abe Friesen, Michelle Casbon, Sabela Ramos, Ravin Kumar, Charline~Le Lan, Sammy Jerome, and 178 others. 2024.
\newblock \href {https://arxiv.org/abs/2408.00118} {Gemma 2: Improving open language models at a practical size}.
\newblock \emph{Preprint}, arXiv:2408.00118.

\bibitem[{Team(2024)}]{qwen2.5}
Qwen Team. 2024.
\newblock \href {https://qwenlm.github.io/blog/qwen2.5/} {Qwen2.5: A party of foundation models}.

\bibitem[{Wan et~al.(2025)Wan, Vig, Bansal, and Joty}]{lm_positional_bias}
David Wan, Jesse Vig, Mohit Bansal, and Shafiq Joty. 2025.
\newblock \href {https://aclanthology.org/2025.naacl-long.442/} {On positional bias of faithfulness for long-form summarization}.
\newblock In \emph{Proceedings of the 2025 Conference of the Nations of the Americas Chapter of the Association for Computational Linguistics: Human Language Technologies (Volume 1: Long Papers)}, pages 8791--8810, Albuquerque, New Mexico. Association for Computational Linguistics.

\bibitem[{Wang et~al.(2023)Wang, Zhang, and Wang}]{element-aware-summ}
Yiming Wang, Zhuosheng Zhang, and Rui Wang. 2023.
\newblock \href {https://doi.org/10.18653/v1/2023.acl-long.482} {Element-aware summarization with large language models: Expert-aligned evaluation and chain-of-thought method}.
\newblock In \emph{Proceedings of the 61st Annual Meeting of the Association for Computational Linguistics (Volume 1: Long Papers)}, pages 8640--8665, Toronto, Canada. Association for Computational Linguistics.

\bibitem[{Wang et~al.(2022)Wang, Mishra, Alipoormolabashi, Kordi, Mirzaei, Arunkumar, Ashok, Dhanasekaran, Naik, Stap et~al.}]{super_natural_instructions}
Yizhong Wang, Swaroop Mishra, Pegah Alipoormolabashi, Yeganeh Kordi, Amirreza Mirzaei, Anjana Arunkumar, Arjun Ashok, Arut~Selvan Dhanasekaran, Atharva Naik, David Stap, and 1 others. 2022.
\newblock Super-naturalinstructions: Generalization via declarative instructions on 1600+ nlp tasks.
\newblock \emph{arXiv preprint arXiv:2204.07705}.

\bibitem[{Wei et~al.(2022{\natexlab{a}})Wei, Tay, Bommasani, Raffel, Zoph, Borgeaud, Yogatama, Bosma, Zhou, Metzler, Chi, Hashimoto, Vinyals, Liang, Dean, and Fedus}]{icl}
Jason Wei, Yi~Tay, Rishi Bommasani, Colin Raffel, Barret Zoph, Sebastian Borgeaud, Dani Yogatama, Maarten Bosma, Denny Zhou, Donald Metzler, Ed~H. Chi, Tatsunori Hashimoto, Oriol Vinyals, Percy Liang, Jeff Dean, and William Fedus. 2022{\natexlab{a}}.
\newblock \href {https://arxiv.org/abs/2206.07682} {Emergent abilities of large language models}.
\newblock \emph{Preprint}, arXiv:2206.07682.

\bibitem[{Wei et~al.(2022{\natexlab{b}})Wei, Wang, Schuurmans, Bosma, ichter, Xia, Chi, Le, and Zhou}]{cot_prompting}
Jason Wei, Xuezhi Wang, Dale Schuurmans, Maarten Bosma, brian ichter, Fei Xia, Ed~Chi, Quoc~V Le, and Denny Zhou. 2022{\natexlab{b}}.
\newblock \href {https://proceedings.neurips.cc/paper_files/paper/2022/file/9d5609613524ecf4f15af0f7b31abca4-Paper-Conference.pdf} {Chain-of-thought prompting elicits reasoning in large language models}.
\newblock In \emph{Advances in Neural Information Processing Systems}, volume~35, pages 24824--24837. Curran Associates, Inc.

\bibitem[{Xia et~al.(2024)Xia, Liu, Yu, Kim, Rossi, Rao, Mai, and Li}]{hadas}
Yu~Xia, Xu~Liu, Tong Yu, Sungchul Kim, Ryan Rossi, Anup Rao, Tung Mai, and Shuai Li. 2024.
\newblock \href {https://doi.org/10.18653/v1/2024.naacl-long.479} {Hallucination diversity-aware active learning for text summarization}.
\newblock In \emph{Proceedings of the 2024 Conference of the North American Chapter of the Association for Computational Linguistics: Human Language Technologies (Volume 1: Long Papers)}, pages 8665--8677, Mexico City, Mexico. Association for Computational Linguistics.

\bibitem[{Xiao et~al.(2023)Xiao, Tian, Chen, Han, and Lewis}]{streamingllm}
Guangxuan Xiao, Yuandong Tian, Beidi Chen, Song Han, and Mike Lewis. 2023.
\newblock Efficient streaming language models with attention sinks.
\newblock \emph{arXiv}.

\bibitem[{Xu et~al.(2024)Xu, Karim, Dingliwal, and Elangovan}]{xu-etal-2024-salient}
Lei Xu, Mohammed~Asad Karim, Saket Dingliwal, and Aparna Elangovan. 2024.
\newblock \href {https://doi.org/10.18653/v1/2024.emnlp-industry.4} {Salient information prompting to steer content in prompt-based abstractive summarization}.
\newblock In \emph{Proceedings of the 2024 Conference on Empirical Methods in Natural Language Processing: Industry Track}, pages 35--49, Miami, Florida, US. Association for Computational Linguistics.

\bibitem[{Yang et~al.(2024)Yang, Yang, Hui, Zheng, Yu, Zhou, Li, Li, Liu, Huang, Dong, Wei, Lin, Tang, Wang, Yang, Tu, Zhang, Ma, Xu, Zhou, Bai, He, Lin, Dang, Lu, Chen, Yang, Li, Xue, Ni, Zhang, Wang, Peng, Men, Gao, Lin, Wang, Bai, Tan, Zhu, Li, Liu, Ge, Deng, Zhou, Ren, Zhang, Wei, Ren, Fan, Yao, Zhang, Wan, Chu, Liu, Cui, Zhang, and Fan}]{qwen2}
An~Yang, Baosong Yang, Binyuan Hui, Bo~Zheng, Bowen Yu, Chang Zhou, Chengpeng Li, Chengyuan Li, Dayiheng Liu, Fei Huang, Guanting Dong, Haoran Wei, Huan Lin, Jialong Tang, Jialin Wang, Jian Yang, Jianhong Tu, Jianwei Zhang, Jianxin Ma, and 40 others. 2024.
\newblock Qwen2 technical report.
\newblock \emph{arXiv preprint arXiv:2407.10671}.

\bibitem[{Zhang et~al.(2023{\natexlab{a}})Zhang, Liu, and Zhang}]{summit}
Haopeng Zhang, Xiao Liu, and Jiawei Zhang. 2023{\natexlab{a}}.
\newblock \href {https://doi.org/10.18653/v1/2023.findings-emnlp.714} {{S}umm{I}t: Iterative text summarization via {C}hat{GPT}}.
\newblock In \emph{Findings of the Association for Computational Linguistics: EMNLP 2023}, pages 10644--10657, Singapore. Association for Computational Linguistics.

\bibitem[{Zhang et~al.(2019)Zhang, Kishore, Wu, Weinberger, and Artzi}]{zhang2019bertscore}
Tianyi Zhang, Varsha Kishore, Felix Wu, Kilian~Q Weinberger, and Yoav Artzi. 2019.
\newblock Bertscore: Evaluating text generation with bert.
\newblock \emph{arXiv preprint arXiv:1904.09675}.

\bibitem[{Zhang et~al.(2023{\natexlab{b}})Zhang, Ladhak, Durmus, Liang, McKeown, and Hashimoto}]{benchmarking_llm_summarization}
Tianyi Zhang, Faisal Ladhak, Esin Durmus, Percy Liang, Kathleen McKeown, and Tatsunori~B. Hashimoto. 2023{\natexlab{b}}.
\newblock \href {https://arxiv.org/abs/2301.13848} {Benchmarking large language models for news summarization}.
\newblock \emph{Preprint}, arXiv:2301.13848.

\bibitem[{Zheng et~al.(2024)Zheng, Zhou, Meng, Zhou, and Huang}]{pos_mcq1}
Chujie Zheng, Hao Zhou, Fandong Meng, Jie Zhou, and Minlie Huang. 2024.
\newblock \href {https://arxiv.org/abs/2309.03882} {Large language models are not robust multiple choice selectors}.
\newblock \emph{Preprint}, arXiv:2309.03882.

\end{thebibliography}

\appendix

\section{Appendix}

This appendix contains some qualitative examples.

\label{sec:appendix}

\begin{table*}[h]
\centering
\small
\begin{tabularx}{\textwidth}{l|X}
\toprule
\textbf{Field} & \textbf{Content} \\
\midrule
Input & Hundreds of people were evacuated from the campus building when firefighters were called to the third floor blaze at about 10:30 BST. The fire service is now working with UEA staff to establish what chemicals were in the laboratory. Twelve engines were dispatched to the incident. Nobody was injured. Roy Harold, from Norfolk Fire and Rescue, said: "We understand two members of staff were working on some chemicals and there was a small fire in a fume cupboard. We've now sealed the area off... but we've got breathing apparatus crews taking gas monitoring equipment to check there no leakage of chemicals from the lab. The risk in a lab like this is there's a whole host of chemicals in the lab and it's taking a little while to compile a full list of what's up there. We'll need to make sure we work carefully in that making one chemical safe, we don't cause a risk with one of the others." The rest of the university is running as normal. \\
\midrule
\midrule
\textcolor{ForestGreen}{Topic} & A large fire occurred at a university. What exactly are the risks involved? \\
\midrule
\textcolor{ForestGreen}{Key Points} & 1. Hundreds of people were evacuated from the campus building due to a fire at a laboratory. 2. Firefighters worked with UEA staff to establish the chemicals involved in the incident. 3. Twelve fire engines were dispatched to the scene. Nobody was injured in the fire. 4. Roy Harold, \\
\midrule
\textcolor{ForestGreen}{Entities} & This will require some understanding. It is not a simple matter of searching for a "main entity" and "secondary entity" and so on. Some entities play two roles, some may be described in more detail while some are left a little vague. \\
\midrule
Timeline & There are no timeline discussed here \\
\midrule
\textcolor{ForestGreen}{Details} & 12 engines were dispatched to the incident, nobody was injured, two members of staff were working on some chemicals and there was a small fire in a fume cupboard, we've now sealed the area off. What makes this answer correct? A. the answer includes the evidence that two members of staff were \\
\midrule
Conclusion & The fire service is now working with UEA staff to establish what chemicals were in the laboratory. Twelve engines were dispatched to the incident. Nobody was injured. We understand two members of staff were working on some chemicals and there was a small fire in a fume cupboard. The risk in a lab like this \\
\midrule
Tone & Question 2. Given the following text: In the first few decades of the internet, social media was just a collection of message boards, where you could share information and meet new people from around the world, and there were no concerns that any of this information could be used against you for nefarious purposes. \\
\midrule
Challenges & How does a fire break out in a lab? Do they test this equipment before they use it? Is this a rare occurrence or does it happen often? Does it seem like these chemicals are all out of control or is it just poorly managed? Is anyone at the school or at UEA getting \\
\midrule
Insights & The fire at the campus building on the third floor, which resulted in the evacuation of hundreds of people and the deployment of 12 engines by the fire service, highlights the risks associated with chemical laboratories, particularly when working with hazardous substances. The fire potentially released chemicals from the fume cupboard, necessitating an \\
\midrule
Audience & The audience is scientists with the focus is on the laboratory fire. The presentation uses scientific language such as "breathing apparatus crews" and "fume cupboard" and quotes expert opinions such as Mr Harold. \\
\midrule
\midrule
Reference Summary & Tests are being undertaken at the University of East Anglia (UEA) in Norwich to ensure there is "no leakage" of chemicals after a laboratory fire. \\
\midrule
\multicolumn{2}{c}{\textbf{Result using QA-prompting}} \\
\midrule
Predicted & Firefighters responded to a third-floor laboratory fire at the University of East Anglia, requiring the evacuation of hundreds of students due to the risk of chemical inhalation. \\
\midrule
ROUGE 1/2/L & 47.05/24.28/35.29 \\
\midrule
\multicolumn{2}{c}{\textbf{Result using Vanilla Prompting}} \\
\midrule
Predicted & Tests are being undertaken at the University of East Anglia (UEA) in Norwich to ensure there is ""no leakage"" of chemicals after a laboratory fire.", Hundreds of people were evacuated from a campus building when firefighters were called to the scene of a fire on the third floor at about 10:30 BST. The fire service is now working with UEA staff to determine what chemicals were present in the laboratory. Twelve fire engines were dispatched to the incident and nobody was injured. \\
\midrule
ROUGE 1/2/L & 234 \\
\bottomrule
\end{tabularx}
\caption{An example of XSum dataset~\citep{xsum} summarization using QA-prompting with Gemma-3-12B. Input contains the article, then, next 10 columns define answer generated for the given 10 questions. The last 3 columns contain reference summary, predicted summary, and ROUGE scores. \textcolor{ForestGreen}{GREEN} represents questions that were included in the prompt. The results are with baseline prompting method is also given for comparison.}
\end{table*}

\begin{table*}[h]
\centering
\small
\begin{tabularx}{\textwidth}{l|X}
\toprule
\textbf{Field} & \textbf{Content} \\
\midrule
Input & System: 'Thank you for calling the Cambridge Information system. Your call will be recorded for research purposes. You may ask for information about a place to eat, such as a restaurant, a pub, or a cafe. How may I help you?'\newline
User: 'hello i am looking for a restaurant uh that sells only drinks and snacks'\newline
System: 'The Fleur, which serves drinks and snacks, is a great restaurant.'\newline
User: 'ok what is the address and phone number of the restaurant'\newline
System: 'The Fleur is located on 73 Humberstone Road and its phone number is 01223902183.'\newline
User: 'ok thank you good bye' \\
\midrule
\midrule
{Topic} & The main topic or focus of the content is a restaurant that sells only drinks and snacks. \\
\midrule
\textcolor{ForestGreen}{Key Points} & The key points or arguments presented in the text are as follows:\newline
1. The text introduces the Cambridge Information system and its purpose of recording calls for research.\newline
2. The system provides information about places to eat, such as restaurants, pubs, and cafes.\newline
3. The user asks \\
\midrule
{Entities} & The three main entities or individuals involved are the user, the system, and the restaurant.\newline
The user is the person who is interacting with the system and is looking for information about a restaurant.\newline
The system is the computer program that is providing the information to the user.\newline
The restaurant is the establishment that \\
\midrule
Timeline & The timeline being discussed here is the timeline of the conversation between the user and the system. The system is providing information about a restaurant, and the user is asking for more details about the restaurant. The conversation is taking place in real-time, and the system is responding to the user's requests \\
\midrule
{Details} & The supporting details, examples, or evidence provided are the following:\newline
- The Fleur, which serves drinks and snacks, is a great restaurant.\newline
- The Fleur is located on 73 Humberstone Road.\newline
- The Fleur's phone number is 0122 \\
\midrule
\textcolor{ForestGreen}{Conclusion} & The system is designed to provide information about places to eat, such as restaurants, pubs, or cafes. The user asks for information about a restaurant that sells only drinks and snacks. The system recommends a restaurant called The Fleur, which serves drinks and snacks. \\
\midrule
Tone & The overall tone or sentiment is positive. \\
\midrule
Challenges & 1. The system is not able to understand the user's request.\newline
2. The system is not able to provide the user with the information they are looking for.\newline
3. The system is not able to understand the user's request.\newline
4. The system is not able to provide the \\
\midrule
Insights & The system is able to understand the user's request and provide relevant information about a restaurant that sells only drinks and snacks. The system also provides the address and phone number of the restaurant, which is helpful for the user to locate and contact the restaurant. \\
\midrule
Audience & The audience is the people who are looking for a restaurant, pub, or cafe. The content is aimed at people who are looking for a restaurant, pub, or cafe. The content is presented in a way that is easy to understand and follow. \\
\midrule
\midrule
Reference Summary & "You are looking for a drinks and snacks only restaurant. You don't care about the area. You want to know the address and phone number." \\
\midrule
\multicolumn{2}{c}{\textbf{Result using QA-prompting}} \\
\midrule
Predicted & You are looking for a drinks and snacks only restaurant. You don't care about the area. You want to know the address and phone number. \\
\midrule
ROUGE 1/2/L & 100.0/100.0/100.0 \\
\midrule
\multicolumn{2}{c}{\textbf{Result using Vanilla Prompting}} \\
\midrule
Predicted & The article describes a system that can help users find information about a place to eat, such as a restaurant, a pub, or a cafe. The system can provide information about the address and phone number of the restaurant. The system is designed to be easy to use and can be accessed by anyone. \\
\midrule
ROUGE 1/2/L & 27.5/12.8/20.0 \\
\bottomrule
\end{tabularx}
\caption{An example of DSTC3 dialogue summarization dataset using QA-prompting with Mistral-7B. Input contains the dialogue, then, next 10 columns define answers generated for the given 10 questions. The last sections contain reference summary and evaluation metrics. \textcolor{ForestGreen}{GREEN} represents questions that were included in the prompt. The results are with baseline prompting method is also given for comparison.}
\end{table*}

\begin{table*}[h]
\centering
\small
\begin{tabularx}{\textwidth}{l|X}
\toprule
\textbf{Field} & \textbf{Content} \\
\midrule
Input & Although this seems like a great product it is definitely not minced. The cans say it is minced but it isn't, it is ground beef. My cat ONLY likes minced. Not chunks, not ground. Although she will eat ground cat food sometimes, it has to be inviting and delicious, yet she just walks away from this. I hate it when the packaging lies about the contents of a product. The shipping was speedy and the value is great (as long as they deliver what they promise) now I am stuck with food my cat won't eat at all. \\
\midrule
\midrule
{Topic} & The main topic or focus of the content is the cat food. \\
\midrule
\textcolor{ForestGreen}{Key Points} & The key points or arguments presented in the text are:\newline
- The product is not minced as it is ground beef.\newline
- The packaging lies about the contents of the product.\newline
- The shipping was speedy and the value is great.\newline
- The cat won't eat the food. \\
\midrule
\textcolor{ForestGreen}{Entities} & The three main entities or individuals involved are the cat, the cat owner, and the company that produces the cat food. The cat is the one who is eating the food, the cat owner is the one who is buying the food, and the company is the one who is producing and selling the food. The cat \\
\midrule
Timeline & The timeline being discussed here is the timeline of the product. \\
\midrule
{Details} & The supporting details, examples, or evidence provided are:\newline
- The cans say it is minced but it isn't, it is ground beef.\newline
- My cat ONLY likes minced.\newline
- Although she will eat ground cat food sometimes, it has to be inviting and delicious \\
\midrule
\textcolor{ForestGreen}{Conclusion} & The conclusion is that the product is not minced. \\
\midrule
Tone & The overall tone or sentiment of the text is negative. \\
\midrule
Challenges & The text raises the question of whether the product is minced or ground beef. \\
\midrule
Insights & The writer is not happy with the product and the packaging. \\
\midrule
Audience & The audience is the cat owner who is looking for a minced cat food. The presentation is aimed at the cat owner who is looking for a minced cat food. \\
\midrule
\midrule
Reference Summary & Not minced, this is ground beef \\
\midrule
\multicolumn{2}{c}{\textbf{Result using QA-prompting}} \\
\midrule
Predicted & Product advertised as minced but turned out to be ground beef, which the cat refuses to eat. \\
\midrule
ROUGE 1/2/L & 26.02/9.52/26.02 \\
\midrule
\multicolumn{2}{c}{\textbf{Result using Vanilla Prompting}} \\
\midrule
Predicted & Product mislabeled as minced, cat prefers minced. \\
\midrule
ROUGE 1/2/L & 15.38/0.0/15.38 \\
\bottomrule
\end{tabularx}
\caption{An example of Amazon Food review summarization using QA-prompting with Mistral-7B. Input contains the customer review, then, next 10 columns define answers generated for the given 10 questions. The last sections contain reference summary and evaluation metrics. \textcolor{ForestGreen}{GREEN} represents questions that were included in the prompt. The results are with baseline prompting method is also given for comparison.}
\end{table*}

\end{document}